\documentclass[10pt,twocolumn,letterpaper]{article}
\pdfoutput=1

\usepackage[table]{xcolor}
\usepackage[pagenumbers]{cvpr}              
\makeatletter
\@namedef{ver@everyshi.sty}{}
\makeatother
\usepackage{tikz}

\usepackage{graphicx}
\usepackage{amsmath}
\usepackage{amssymb}
\usepackage{microtype}

\usepackage{enumitem}
\setlist{leftmargin = *} 

\usepackage{booktabs}
\usepackage{tabularx}
\usepackage{csvsimple}
\usepackage{siunitx}
\usepackage{xspace}

\renewcommand{\paragraph}[1]{\noindent\textbf{#1} \quad}

\usepackage{color}
\newcommand{\jf}[1]{{\color{red}{[JF: #1]}}}
\newcommand{\yohan}[1]{{\color{orange}{[Yohan: #1]}}}

\renewcommand{\jf}[1]{}
\renewcommand{\yohan}[1]{}

\usepackage[linesnumbered,ruled,vlined]{algorithm2e}

\SetKwComment{Comment}{\color{green!50!black}\# }{}

\newcommand{\assign}{\leftarrow}

\SetKwProg{Function}{function}{}{}

\DontPrintSemicolon

\DeclareMathOperator*{\argmin}{arg\,min}

\renewcommand{\etal}{et al.\xspace}
\renewcommand{\eg}{e.g.\xspace}
\renewcommand{\ie}{i.e.\xspace}

\definecolor{best}{RGB}{255, 220, 200}
\definecolor{second}{RGB}{255, 255, 200}

\usepackage{hyperref}
\hypersetup{pagebackref,breaklinks,colorlinks}

\usepackage[capitalize]{cleveref}
\crefname{algorithm}{alg.}{algs.}
\Crefname{algorithm}{Alg.}{Algs.}
\crefname{figure}{fig.}{figs.}
\Crefname{figure}{Fig.}{Figs.}
\crefname{section}{sec.}{secs.}
\Crefname{section}{Sec.}{Secs.}
\crefname{equation}{eq.}{eqs.}
\Crefname{equation}{Eq.}{Eqs.}
\crefname{table}{tab.}{tabs.}
\Crefname{table}{Tab.}{Tabs.}

\def\cvprPaperID{8084} 
\def\confName{CVPR}
\def\confYear{2023}

\usepackage[utf8]{inputenc}
\usepackage[margin=1in,bottom=1in,top=1in]{geometry}
\usepackage{csvsimple}

\begin{document}
\title{Robust Unsupervised StyleGAN Image Restoration}


\author{Yohan Poirier-Ginter$^{\bullet\diamond}$, 
Jean-Fran\c{c}ois Lalonde$^{\bullet}$\\
$^\diamond$Inria, Université Côte d'Azur, $^\bullet$Université Laval \\ 
\small{\texttt{\url{https://lvsn.github.io/RobustUnsupervised/}}} 
} 

\twocolumn[{%
\renewcommand\twocolumn[1][]{#1}%

\maketitle
\vspace{-10mm}
\begin{center}
    \centering
    \captionsetup{type=figure}
    \includegraphics[width=\textwidth]{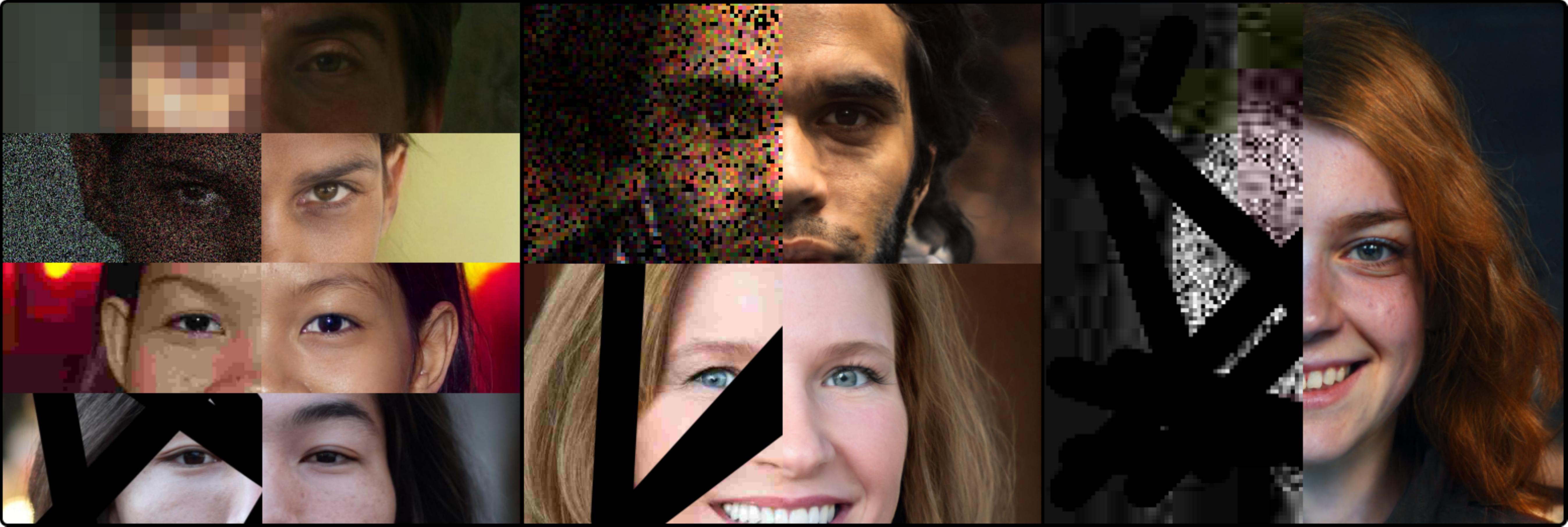}
    \vspace{-1.5em}
    \captionof{figure}{Our unsupervised StyleGAN image restoration method is robust and effective across a variety of tasks (left, top to bottom: upsampling, denoising, deartifacting, and inpainting) and a wide range of degradation levels. Since it avoids the need for task-specific hyperparameter tuning, it can directly handle \emph{combinations} of such degradations, \eg, pairs (center) or even all four (right).}
    \label{fig:teaser}
\end{center}%
}]

\begin{abstract}
GAN-based image restoration inverts the generative process to repair images corrupted by known degradations. Existing unsupervised methods must be carefully tuned for each task and degradation level. In this work, we make StyleGAN image restoration robust: a single set of hyperparameters works across a wide range of degradation levels. This makes it possible to handle combinations of several degradations, without the need to retune. Our proposed approach relies on a 3-phase progressive latent space extension and a conservative optimizer, which avoids the need for any additional regularization terms. Extensive experiments demonstrate robustness on inpainting, upsampling, denoising, and deartifacting at varying degradations levels, outperforming other StyleGAN-based inversion techniques. Our approach also favorably compares to diffusion-based restoration by yielding much more realistic inversion results. Code is available at the above URL.
\end{abstract}

\vspace{-4mm}
\section{Introduction}
\label{sec:intro}

Image restoration, the task of recovering a high quality image from a degraded input, is a long-standing problem in image processing and computer vision. Since different restoration tasks---such as denoising, upsampling, deartifacting, etc.---can be quite distinct, many recent approaches~\cite{chen2021pre,chen2020proxiqa,liang2021swinir,zamir2020learning,zamir2021multi,tu2022maxim} propose to solve them in a supervised learning paradigm by leveraging curated datasets specifically designed for the task at hand. Unfortunately, designing task-specific approaches requires retraining large networks on each task separately.

In parallel, the advent of powerful generative models has enabled the emergence of \emph{unsupervised} restoration methods~\cite{exploiting}, which do not require task-specific training. The idea is to invert the generative process to recover a clean image. Assuming a known (or approximate) degradation model, the optimization procedure therefore attempts to recover an image that both: 1) closely matches the target degraded image after undergoing a similar degradation model (fidelity); and 2) lies in the space of realistic images learned by the GAN (realism). 

In the recent literature, StyleGAN~\cite{stylegan,stylegan2,stylegan3} has been found to be particularly effective for unsupervised image restoration~\cite{pulse, brgm, ilo, l-brgm, sgilo} because of the elegant design of its latent space. Indeed, these approaches leverage style inversion techniques~\cite{image2stylegan,image2stylegan++} to solve for a latent vector that, when given to the generator, creates an image close to the degraded target. Unfortunately, this only works when such a match actually exists in the model distribution, which is rarely the case in practice. Hence, effective methods \emph{extend} the learned latent space to create additional degrees of freedom to admit more images; this creates the need for additional regularization losses.
Hyperparameters must therefore carefully be tuned for each specific task and degradation level. 

In this work, we make unsupervised StyleGAN image restoration-by-inversion \emph{robust} to the type and intensity of degradations. 
Our proposed method employs the \emph{same} hyperparameters across all tasks and levels and does not rely on any regularization loss. Our approach leans on two key ideas. First, we rely on a 3-phase \emph{progressive latent space extension}: we begin by optimizing over the learned (global) latent space, then expand it across individual \emph{layers} of the generator, and finally expand it further across individual \emph{filters}---where optimization at each phase is initialized with the result of the previous one. Second, we rely on a conservative, normalized gradient descent (NGD)~\cite{ml-refined} optimizer which is naturally constrained to stay close to its initial point compared to more sophisticated approaches such as Adam~\cite{adam}. 
This combination of prudent optimization over a progressively richer latent space avoids additional regularization terms altogether and keep the overall procedure simple and constant across all tasks. We evaluate our method on upsampling, inpainting, denoising and deartifacting on a wide range of degradation levels, achieving state-of-the-art results on most scenarios even when baselines are optimized on each independently. We also show that our approach outperforms existing techniques on \emph{compositions} of these tasks without changing hyperparameters.

\begin{itemize}[noitemsep,topsep=0pt]
    \item We propose a robust 3-phase StyleGAN image restoration framework. Our optimization technique maintains: 1) strong realism when degradations level are high; and 2) high fidelity when they are low. Our method is fully unsupervised, requires no per-task training, and can handle different tasks at different levels without having to adjust hyperparameters.
    \item We demonstrate the effectiveness of the proposed method under \emph{diverse} and \emph{composed} degradations. We develop a benchmark of synthetic image restoration tasks---making their degradation levels easy to control---with care taken to avoid unrealistic assumptions. Our method outperforms existing unsupervised~\cite{pulse,l-brgm} and diffusion-based~\cite{ddrm} approaches.
\end{itemize}

\section{Related work}
\label{sec:related-work}

This section covers previous work on StyleGAN inversion by optimization and its use for image restoration, and discusses related works on generative priors. Because the literature on general purpose image restoration is so extensive, we do not attempt to review it here and instead refer the reader to recent surveys~\cite{superres-survery,superres-survey-2,inpainting-review,denoising-review}.

\paragraph{GANs and StyleGAN inversion} 
\label{sec:stylegan-inversion}
Generative adversarial networks~\cite{gan} (GANs) are a popular technique for generative modeling. In particular, the StyleGAN family~\cite{stylegan, stylegan2, stylegan2-ada, stylegan3} learns a highly compressed latent representation of its domain and stands out for its popularity and generation quality. Great progress has been made in StyleGAN \emph{inversion}: reversing the generation process to infer latent parameters that generate a given image~\cite{inversion-survey}. 

Thanks to inversion, StyleGAN has found use in many image editing~\cite{interfacegan, ganspace, styleclip} and restoration~\cite{pulse, ilo, sgilo, brgm, l-brgm} tasks. Purely optimization-based techniques were studied extensively~\cite{image2stylegan, image2stylegan++, bdinvert, overparam, where-are-good-latents, pivotal-tuning} in the context of image editing. These methods extend the learned latent space of a pretrained StyleGAN model (commonly named $\mathcal{W}$) by adding additional parameters to optimize over it. The most common approach consists of using a different latent code for each layer~\cite{image2stylegan} (dubbed $\mathcal{W}^+$). Going beyond $\mathcal{W}^+$ has also been explored in \cite{pivotal-tuning} who propose to fine-tune generator parameters, and \cite{overparam} which uses different latent codes for each convolution filter. We build on these techniques by developing an inversion method designed specifically for robust image restoration.

\paragraph{Generative priors for unsupervised image restoration}
    Since the initial work of Bora~\etal~\cite{compressed-sensing}, multiple papers have used generative priors for unsupervised image restoration~\cite{image-adaptive-gan, mgan-prior}. Pan~\etal~\cite{exploiting} obtained high-resolution results with BigGAN~\cite{biggan}. They relax their latent inversion method by introducing a coarse-to-fine, layer-by-layer generator fine-tuning. Our method differs in that it addresses robustness and compositionality. 

\paragraph{StyleGAN image restoration} Inversion methods designed for editing cannot be applied directly for restoration because loss functions are applied to degraded targets rather than clean images. Previous works address this with additional regularization terms and/or optimizer constraints. Of note, PULSE~\cite{pulse} proposed StyleGAN inversion for image upsampling, using a regularizer minimizing the latent extension to $\mathcal{W}^+$ and a spherical optimization technique on a Gaussian approximation of $\mathcal{W}$. While PULSE succeeds in preserving high realism, it obtains low fidelity when the downsampling factor is small.
ILO~\cite{ilo} uses similar regularization techniques, and further extends the latent space by optimizing intermediate feature maps in a progressive fashion (layer-by-layer) while constraining the solutions to sparse deviations from the range. More recently, SGILO~\cite{sgilo} proposes to replace this sparsity constraint and instead trains a score-based model to learn the distribution of the outputs of an intermediate layer in the StyleGAN generator. 
BRGM~\cite{brgm} frames GAN inversion as Bayesian inference and estimates the maximum a-posteriori over the input latent vector that generated the reconstructed image given different regularization constraints. Its followup, L-BRGM~\cite{l-brgm}, jointly optimizes the $\mathcal{Z}$ (before the mapping network) and $\mathcal{W}$ spaces, further improving quality. In contrast, our work achieves robustness by avoiding any such regularization losses.


\paragraph{Other similar works} StyleGAN inversion was used to restore old photographs in \cite{time-travel-rephoto}. Their method also restores a mixture of degradations, but is designed for real degradations (not robustness) and uses a task-specific encoder, while ours method is fully unsupervised. Multiple supervised methods~\cite{ebf, ffsb, blind-fr, GFPGAN, zhou2022towards, gu2022vqfr} focus on faces only, whereas our method does not use face-specific loss functions~\cite{arcface}, is fully unsupervised, and works across a variety of image domains. 

\paragraph{Diffusion models}
Denoising diffusion probabilistic models (DDPMs)~\cite{ddpm}, trained on extremely large datasets (\eg, billions of images for \cite{dalle2, imagen, stable-diffusion}), have recently been shown to outclass GANs when it comes to image generation quality~\cite{diffusion-beat-gans}. New approaches have been proposed to tackle image restoration using DDPMs, \eg, Palette~\cite{palette} obtains excellent results by training the same model for several different tasks. Denoising diffusion restoration models (DDRM)~\cite{ddrm} show that pretrained DDPMs can be used for unsupervised restoration tasks, but are limited to linear inverse problems with additive Gaussian noise. In contrast, our method is more flexible since it only requires a differentiable approximation of the degradation function, which can be non-linear. 
\begin{figure}[t]
    \centering
    \includegraphics[width=\linewidth]{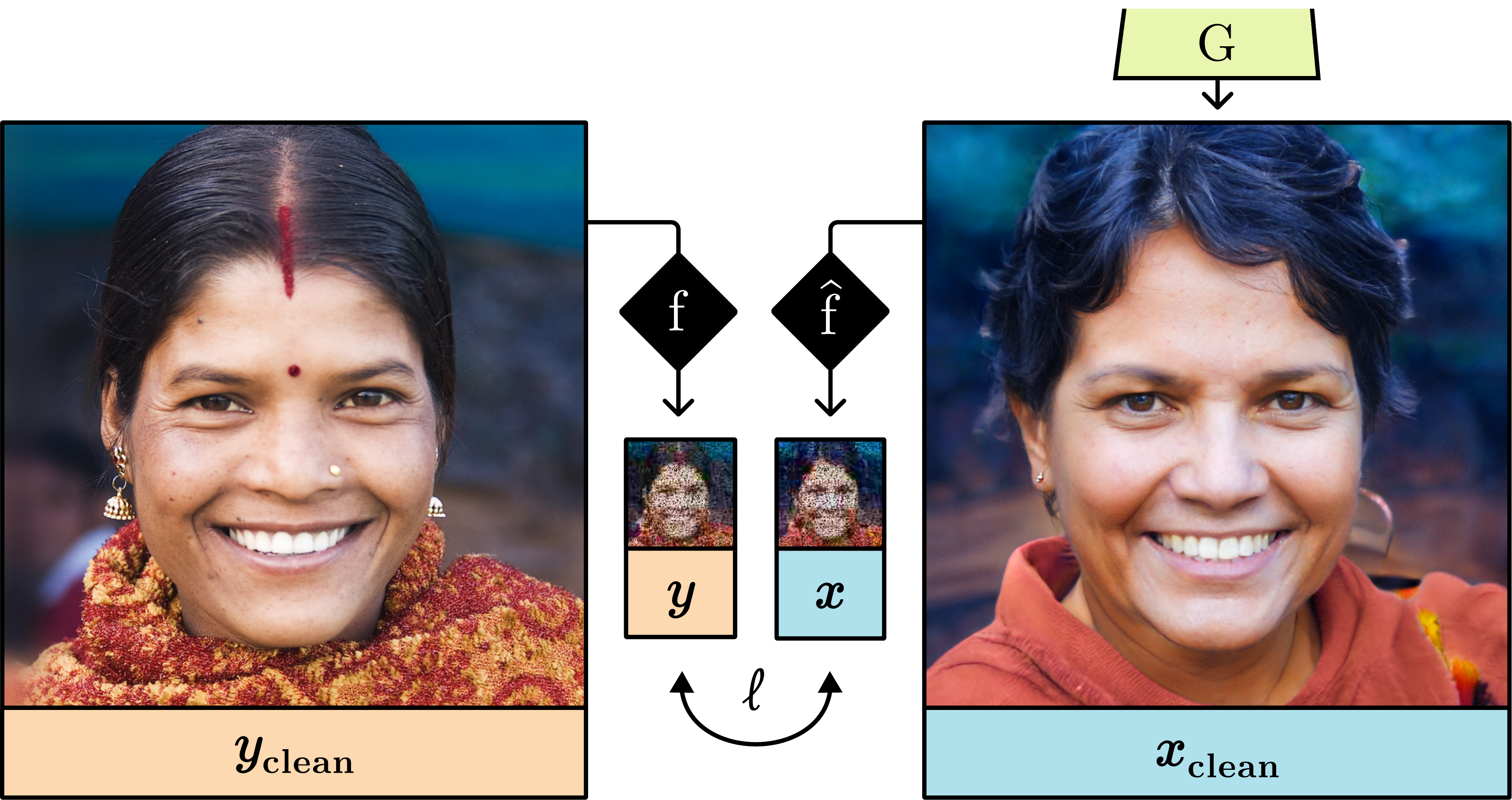}
    \caption{
    Unsupervised image restoration with StyleGAN attempts to restore a degraded target image $\boldsymbol{y} = f(\boldsymbol{y}_\mathrm{clean})$ by finding a generated image $\boldsymbol{x}_\mathrm{clean}$ that matches this target once also degraded $\boldsymbol{x} = \hat{f}(\boldsymbol{x}_\mathrm{clean})$ (sec.~\ref{sec:image-restoration-with-stylegan}); here, $\hat{f}$ is a differentiable approximation to the degradation function $f$ and $\boldsymbol{y}_\mathrm{clean}$ is the (unknown) ground truth image. }
    \label{fig:method-overview}
\end{figure}

\section{Image restoration with StyleGAN}
\label{sec:image-restoration-with-stylegan}

As illustrated in \cref{fig:method-overview}, StyleGAN inversion attemps to recover an image $\boldsymbol{x}_\mathrm{clean}$ that best matches an (unknown) ground truth image $\boldsymbol{y}_\mathrm{clean}$. To this end, we aim to search for a latent code $\boldsymbol{w} \in \mathcal{W}$ such that $\boldsymbol{x}_\mathrm{clean} = G(\boldsymbol{w})$ best matches $\boldsymbol{y}_\mathrm{clean}$ under some image distance function $\ell$. The resulting minimization problem, 
\begin{equation}
\label{eq:min-gt}
\boldsymbol{w} = \argmin_{\tilde{\boldsymbol{w}} \in \mathcal{W}}  \ell\big(G(\tilde{\boldsymbol{w}}), \boldsymbol{y}_\mathrm{clean}\big) \,,    
\end{equation}
can be solved by gradient descent. In image restoration, the ground truth image $\boldsymbol{y}_\mathrm{clean}$ is unknown: we are instead given a target image $\boldsymbol{y} = f(\boldsymbol{y}_\mathrm{clean})$, the result of a (non-injective, potentially non-differentiable) degradation function $f$. Assuming that a differentiable approximation $\hat{f} \approx f$ can be constructed, restoration is performed by solving for
\begin{equation}
\label{eq:min-degraded}
\boldsymbol{w} = \argmin_{\tilde{\boldsymbol{w}} \in \mathcal{W}} \ell\big(\hat{f}(G(\tilde{\boldsymbol{w}})), \boldsymbol{y}\big) \,.
\end{equation} 

This can be generalized to compositions of $k$ different degradations, \ie, $\{f_i\}_{i=1}^k$, by solving for
\begin{equation}
\label{eq:min-composed}
\boldsymbol{w} = \argmin_{\tilde{\boldsymbol{w}} \in \mathcal{W}} \ell\big([\hat{f}_k \circ ... \circ \hat{f}_2 \circ \hat{f}_1](G(\tilde{\boldsymbol{w}})), \boldsymbol{y}\big) \,,
\end{equation}
where $\circ$ is the composition operator. Here, it is assumed that each subfunction $f_i$ has a differentiable approximation $\hat{f}_i$, and that the order of composition is known. 

As described in sec.~\ref{sec:related-work}, this naïve approach finds solutions that have high realism (look like real faces), but low fidelity (do not match the degraded target). Fidelity is most commonly improved by 1) performing \emph{latent extension}~\cite{image2stylegan}, that is, solving for $\boldsymbol{w}^+ \in \mathcal{W}^+$ which has many more degrees of freedom; and 2) using a better performing optimizer like Adam~\cite{adam}. These techniques both improve fidelity but also damage realism, motivating the use of regularization losses which must carefully be adjusted for different tasks~\cite{pulse,brgm,l-brgm}. In the next section, we show how the choice of latent extension and optimizer avoids the need for these regularizers.

\begin{figure}[t]
    \centering
    \footnotesize
    \setlength{\tabcolsep}{1pt}
    \begin{tabular}{ccc}
    \includegraphics[height=5cm,trim=0 0 0 1.8cm,clip]{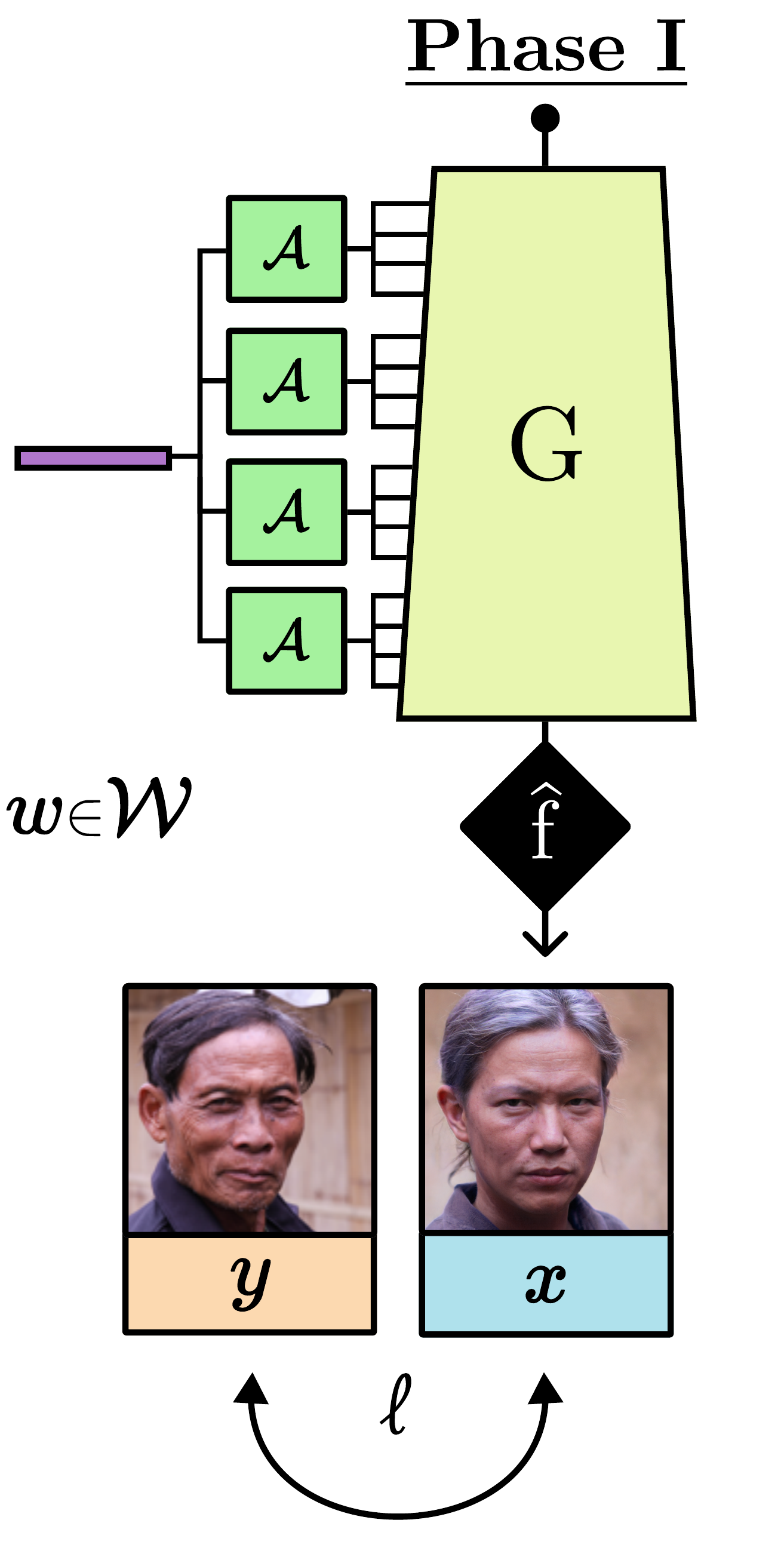} &
    \includegraphics[height=5cm,trim=0 0 0 1.8cm,clip]{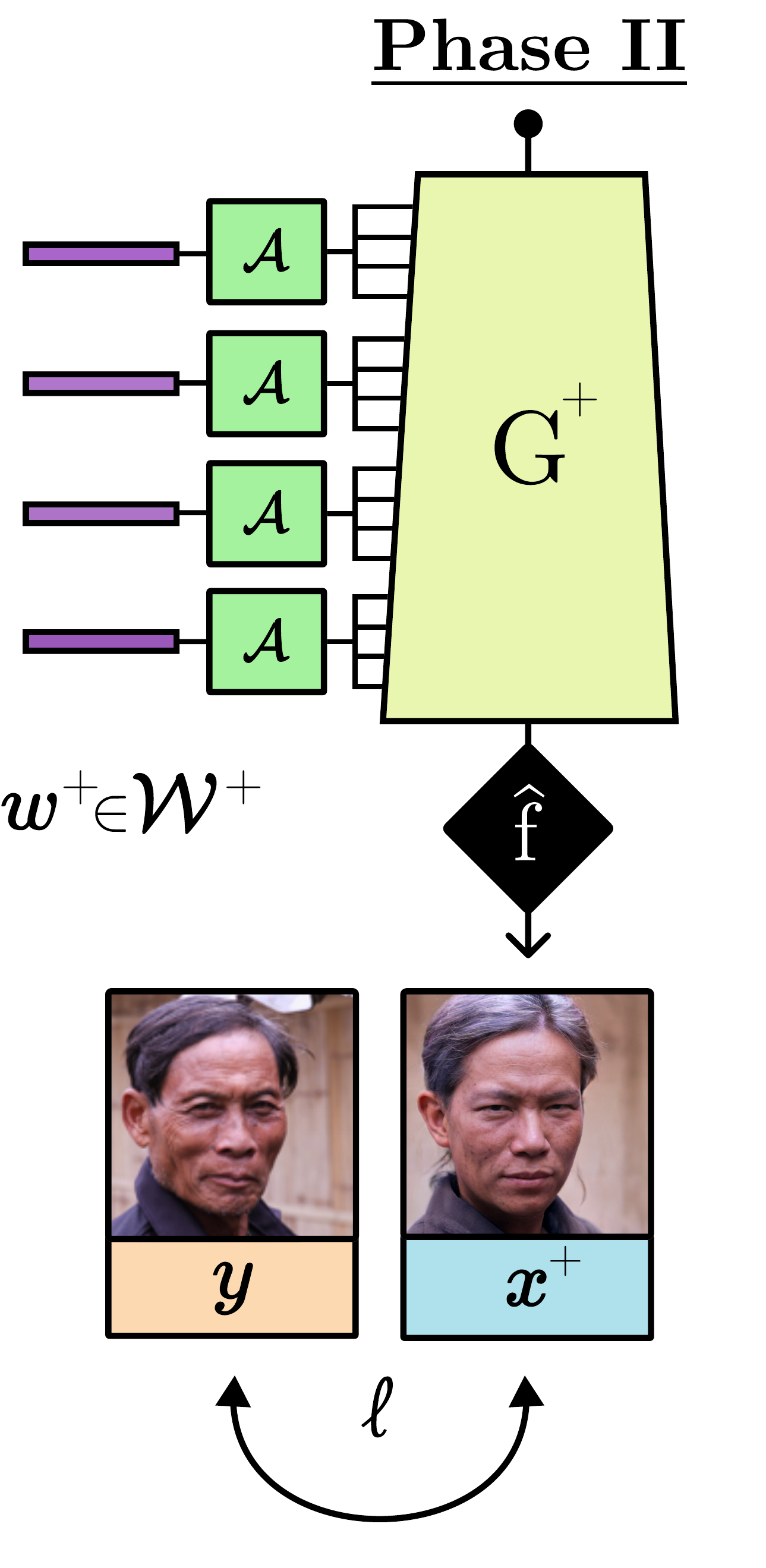} &
    \includegraphics[height=5cm,trim=0 0 0 1.8cm,clip]{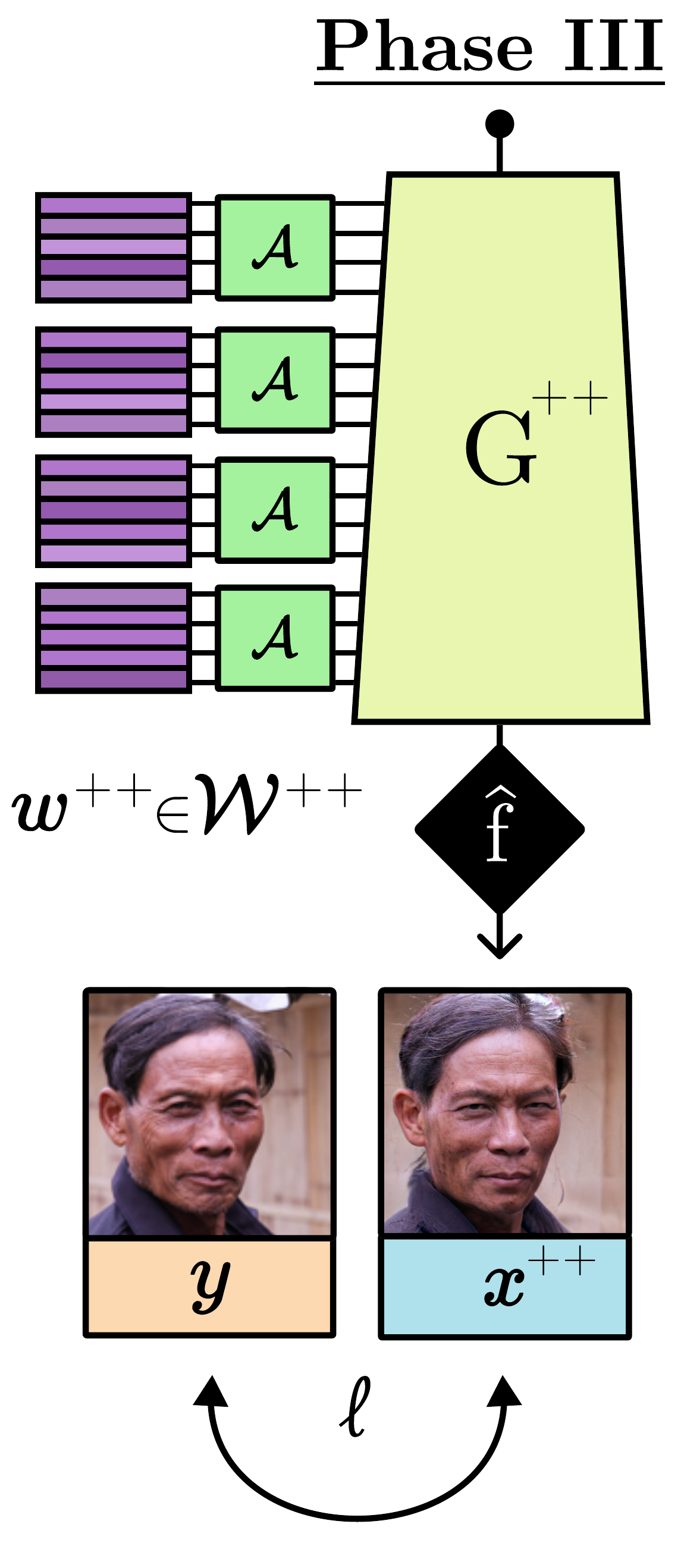} \\
    \small{(a) Phase I} & \small{(b) Phase II} & (c) \small{Phase III}
    \end{tabular}
    \caption{
    Our 3-phase latent extension (\cref{sec:robust-extension}) approach. (a) Phase I uses a \emph{global} latent code $\boldsymbol{w}$, resulting in a prediction $\boldsymbol{x}$. (b) Phase II performs \emph{layer}-wise latent expansion, resulting in matrix $\boldsymbol{w}^{+}$ and in a prediction $\boldsymbol{x}^{+}$. (c) Phase III performs \emph{filter}-wise latent expansion, resulting in tensor $\boldsymbol{w}^{++}$ and the final prediction $\boldsymbol{x}^{++}$.}
    \label{fig:method}
\end{figure}


\section{Robust StyleGAN inversion}
\label{sec:robust-inversion}

We propose an inversion method designed specifically for robust image restoration, which revisits each aspect of the unsupervised StyleGAN optimization pipeline, namely latent extension, optimization, and loss functions.


\subsection{Robust latent extension} 
\label{sec:robust-extension}

Inspired by the intuition that initialization is the best regularization~\cite{pivotal-tuning}, we propose a three-phase latent extension, where each phase is initialized by the outcome of the previous one, see \cref{fig:method}.
Given a pretrained StyleGAN2~\cite{stylegan2} model with $N_L$ layers in the generator, we denote style vectors $\boldsymbol{s}^l_{i} \in \mathbb{R}^{512}$ at layer $l \in [1, N_L]$ used to modulate convolution weights $\boldsymbol{\theta}^l \in \mathbb{R}^{512 \times 512}$. Assuming $1 \times 1$ filters to simplify notation\footnote{In the $3 \times 3$ case, operations are repeated spatially and summed.}, each feature map pixel $\boldsymbol{p}^l \in \mathbb{R}^{512}$ is processed by
\begin{equation}
p^{l+1}_i = (\boldsymbol{\theta}^l_i \odot \boldsymbol{s}_i^l) \boldsymbol{p}^l \,,
\label{eq:style-modulation}
\end{equation}
producing output feature $p^{l+1}_i \in \mathbb{R}$ where $i \in [0, 512]$.


%
%


\paragraph{Phase I} performs \emph{global} style modulation (\cref{fig:method}-(a)), and solves for a single latent vector $\boldsymbol{w} \in \mathcal{W} = \mathbb{R}^{512}$ shared across all layers as in \cref{eq:min-composed}. Prior to optimization, $\boldsymbol{w}$ is initialized to the mean of $\mathcal{W}$ over the training set, \ie, $\mathbb{E}_{\tilde{\boldsymbol{w}} \in \mathcal{W}}[\tilde{\boldsymbol{w}}]$. Here, the style modulation vector $\boldsymbol{s}_i^l$ in \cref{eq:style-modulation} can be written as
\begin{equation}
\boldsymbol{s}_i^l = \mathcal{A}^l(\boldsymbol{w}) \,,
\label{eq:regular-style-injection}
\end{equation}
where $\mathcal{A}^l$ is the corresponding affine projection layer 
(which multiplies by a weight matrix and adds a bias). 

\paragraph{Phase II} performs \emph{layer}-wise latent extension~\cite{image2stylegan} (\cref{fig:method}-(b)), and solves for a latent matrix $\boldsymbol{w}^{+} \in \mathcal{W}^{+}  = \mathbb{R}^{N_{L} \times 512}$. Each row of $\boldsymbol{w}^{+}$ is initialized to $\boldsymbol{w}$. Style modulation for this phase becomes
\begin{equation}
\boldsymbol{s}_i^l = \mathcal{A}^l(\boldsymbol{w}^+_l) \,.
\label{eq:layer-extension}
\end{equation}

\paragraph{Phase III} performs \emph{filter}-wise latent extension (\cref{fig:method}-(c)), and solves for a latent tensor $\boldsymbol{w}^{++} \in \mathcal{W}^{++} = \mathbb{R}^{N_{F} \times N_{L} \times 512}$, where a different latent code is used for each convolution filter and where $N_{F}$ is the number of such filters. Each submatrix of $\boldsymbol{w}^{++}$ is initialized to $\boldsymbol{w}^{+}$. Style modulation for this phase becomes
\begin{equation}
\boldsymbol{s}_i^l = \mathcal{A}^l(\boldsymbol{w}_{i,l}^{++}) \,.
\label{eq:filter-extension}
\end{equation}



\begin{algorithm}[t]
  \caption{Robust StyleGAN inversion. }
  \label{pseudocode}
  \KwOut{restored image $\boldsymbol{x}^{++}$}
  \small\Comment{Phase I}
  $\boldsymbol{w} = \mathbb{E}_{\tilde{\boldsymbol{w}} \in \mathcal{W}}[\tilde{\boldsymbol{w}}]$\;
  \For{1 to 150}{
    $\boldsymbol{x} \assign G(\boldsymbol{w})$\;
    $\boldsymbol{w} \assign 0.08 \bar\nabla_{\boldsymbol{w}}\ell(\hat{f}(\boldsymbol{x}), \boldsymbol{y})$\; 
  }
  \small\Comment{Phase II}
  $\boldsymbol{w}^{+} = \mathrm{repeat}(\boldsymbol{w}, N_L)$\;
  \For{1 to 150}{
    $\boldsymbol{x}^{+} \assign G^{+}(\boldsymbol{w}^{+})$\;
    $\boldsymbol{w}^{+} \assign 0.02 \bar\nabla_{\boldsymbol{w}}\ell(\hat{f}(\boldsymbol{x}^{+}), \boldsymbol{y})$\; 
  }
  \small\Comment{Phase III}
  $\boldsymbol{w}^{++} = \mathrm{repeat}(\boldsymbol{w}^{+}, N_F)$\;
  \For{1 to 150}{
    $\boldsymbol{x}^{++} \assign G^{++}(\boldsymbol{w}^{++})$\;
    $\boldsymbol{w}^{++} \assign 0.005 \bar\nabla_{\boldsymbol{w}^{++}}\ell(\hat{f}(\boldsymbol{x}^{++}), \boldsymbol{y})$\; 
  }
  \Return{$\boldsymbol{x}^{++}$}
\end{algorithm}


\label{method:optimization}
\subsection{Robust optimization} 
While most inversion approaches use the Adam~\cite{adam} optimizer to solve \cref{eq:min-composed}, we instead propose coupling our 3-phase latent extension with a weaker optimizer, normalized gradient decent (NGD)~\cite{ml-refined}, a simple variant of SGD that normalizes the gradient before each step:
\begin{equation}
\label{eq:normalized-gradient}
\bar\nabla_{\boldsymbol{w}}\ell(\boldsymbol{w}) = \frac{\nabla_{\boldsymbol{w}}\ell(\boldsymbol{w})}{||\nabla_{\boldsymbol{w}} \ell(\boldsymbol{w})||_2} \,,
\end{equation} 
Here, the gradient $\bar\nabla_{\boldsymbol{w}}\ell(\boldsymbol{w})$ is explicitly set to 0 when $||\nabla_{\boldsymbol{w}} \ell(\boldsymbol{w})||_2 = 0$. Following latent extension, we normalize each latent code separately (\ie each row of $\boldsymbol{w}^+$ and $\boldsymbol{w}^{++}$). NGD conserves loss scale invariance, a key property of Adam that avoids learning rate adjustments following loss function changes. 

\subsection{Robust loss function} 
\label{method:loss-function} 

The staple loss function in StyleGAN inversion is the LPIPS~\cite{lpips} perceptual loss combined with a L2 or L1~\cite{image2stylegan++} pixel-wise loss. We found a multiresolution loss function to be much more robust, and use:
\begin{equation}
    \label{eq:robust-loss-function}
    \ell_\mathrm{MR}(\boldsymbol{x}, \boldsymbol{y}) = \sum_{i=1}^k \ell_\mathrm{LPIPS}(\phi(\boldsymbol{x}, 2^i), \phi(\boldsymbol{y}, 2^i)) \,,
\end{equation}
where $\phi(\cdot, 2^i)$ downsamples by a factor of $2^i$ using average pooling, and we set $k = 6$ for image resolution of $1024^2$. All resolutions are weighted equally, giving our final loss function $\ell = \lambda_\mathrm{L1} \ell_\mathrm{L1} + \ell_\mathrm{MR}$ with $\lambda_\mathrm{L1} = 0.1$. 


\subsection{Overall algorithm}
Alg.~\ref{pseudocode} provides a detailed pseudo-code of our method. $G^+$ and $G^{++}$ denote the synthesis network $G$ after modification to accept $\boldsymbol{w}^{+} \in \mathcal{W}^{+}$ and $\boldsymbol{w}^{++} \in \mathcal{W}^{++}$, respectively. Hyperparameters like learning rates and number of steps are explicitly provided since they are held constant across all tasks. Their values were found by cross-validation on a subset of the FFHQ training set.

\section{Benchmarking restoration robustness}
\label{sec:benchmarking}

This section first describes the proposed degradation models used in all experiments as well as their respective differentiable approximations. It then explains how compositions of tasks were formed. 


\subsection{Individual degradations}
\label{sec:individual-degradations}

Experiments are performed on four common sources of image degradation: upsampling, inpainting, denoising, and deartifacting. Synthetic models are used to facilitate comparison at different degradation levels. Each degradation is tested at five levels, which are referred to as \textit{extra-small} (XS), \textit{small} (S), \textit{medium} (M), \textit{large} (L), and \textit{extra-large} (XL). Parameters for each degradation level will be provided below for each task in that same order. \Cref{fig:single-degradations} shows examples of all four degradations at the XS and XL levels, see supp.

\paragraph{Upsampling} targets $\boldsymbol{y}$ are produced by downsampling ground truth images $\boldsymbol{y}_\mathrm{clean}$ by integer factors $k_\mathrm{down} \in \{2, 4, 8, 16, 32\}$. Downsampling filters are uniformly sampled from the commonly used bilinear, bicubic, and Lanczos filters, which provides a coarse but broad range of aliasing profiles. During inversion, average pooling is used as the approximation $\hat{f}$ in all cases. 

\paragraph{Inpainting} aims to predict missing regions in an image. To test a variety of masking conditions, random masks are generated by drawing $k_\mathrm{strokes} \in \{1, 5, 9, 13, 17\}$ random strokes of width $0.08 r$ where $r$ is the image resolution, each stroke connecting two random points situated in the outer thirds of the image. In this task, $\hat{f} = f$ \ie we assume that the mask is known. To avoid the bias introduced by black pixels when evaluating the LPIPS loss, identical noise of distribution $\mathcal{N}(0.5, 1)$ is added to the masked region of both the prediction and the target before computing LPIPS. 

\paragraph{Denoising} targets are generated by using a mixture of Poisson and Bernouilli noise, simulating common sources of noise in cameras, namely shot noise and dead (or hot) pixels respectively. This is a challenging scenario because unlike Gaussian noise, Poisson noise is non-additive and signal-dependent, while Bernouilli noise is biased. Additionally, both are non-differentiable. We used the parameters
$k_\mathrm{p} \in \{96, 48, 24, 12, 6\}$ and $k_\mathrm{b} \in \{0.04, 0.08, 0.16, 0.32, 0.64\}$ where
$k_\mathrm{p}$ gives the most likely value added to a pixel (independently for each channel) according to a Poisson distribution, and $k_\mathrm{b}$ the probability that all channels of a pixel are replaced with black. We also account for clamping during image serialization, resulting in the overall noise model:
\begin{equation}
\frac{\mathrm{clamp}(p' \cdot m)}{255}, \text{where} \, p' \sim \mathcal{P}(k_\mathrm{p} p) ,  m \sim \mathcal{B}(k_\mathrm{b}) \,,
\label{eq:noise-model}
\end{equation}
for a ground truth image pixel value $p \in [0.0, 1.0]$ and where $\mathrm{clamp}(\cdot)$ saturates all values outside of $[0, 255]$. Note that $p'$ can only take discrete values. For the differentiable approximation $\hat{f}$, we replace the (non-differentiable, discrete) Poisson noise with a Gaussian approximation $\mathcal{N}(k_\mathrm{p} p - 0.5, \sqrt{k_\mathrm{p} p})$ and treat the Bernoulli noise as an unknown mask. We also use a surrogate gradient for $\mathrm{clamp}$:
\begin{equation}
\frac{d}{dp}\mathrm{clamp}(p) \approx \frac{d}{dp}\mathrm{sigmoid}(2 \cdot (p - 0.5)) \,.
\label{eq:clamp-diff}
\end{equation}

\paragraph{Deartifacting} is performed on JPEG images compressed with \textit{libjpeg}~\cite{libjpeg, clark2015pillow} at quality levels $k_\mathrm{jpeg} \in \{18, 15, 12, 9, 6\}$. The lossy part of JPEG compression can be expressed as
\begin{equation}
\mathrm{quantize}\circ\mathrm{DCT}\circ\mathrm{to\_blocks}\circ\mathrm{chroma\_subsample}\,.
\end{equation}
Here, 4:2:2 chroma subsampling is used. 
This final quantization step rounds each DCT component $c^{(k)}_{i,j}$ positioned in the $k^\text{th}$ block at indices $i, j$ with
\begin{equation}
\mathrm{quantize}(c^{(k)}_{i, j}) = \lfloor c^{(k)}_{i, j} / Q_{i, j} \rceil \,,
\end{equation}
where $Q_{i, j} \in [1, 255] \subset \mathbb{N}$ is the corresponding value from the quantization table. Rounding is the only non-differentiable part of this chain (the lossless part of JPEG compression does not need to be differentiated). We read the quantization tables from file headers and use Lomnitz's implementation~\cite{diffjpeg} of \cite{jpeg-resistant}, which proposes differentiable rounding:
\begin{equation}
\lfloor p \rceil \approx \lfloor p \rceil + (p - \lfloor p \rceil)^3 \,.
\end{equation}
We use this approximation to define a surrogate gradient (keeping the forward pass exact), and interpolate with a straight-through gradient estimator at $\alpha=0.8$:
\begin{equation}
\frac{d}{dp} \lfloor p \rceil \approx \frac{d}{dp} \Big[(1 - \alpha) p + \alpha (p - \lfloor p \rceil)^3  \Big] \,.
\end{equation}

\subsection{Composed degradations}
\label{composed-degrdations}

\Cref{eq:min-composed} assumes knowledge of the order in which degradations are applied. In this work, the following restoration order is used: 
\begin{equation}
\mathrm{inpaint} \circ \mathrm{deartifact} \circ \mathrm{denoise} \circ \mathrm{upsample} \,.
\label{eq:composition-order}
\end{equation}
Task compositions are created with (subsequences of) this ordering. For example, upsampling and deartifacting form a composition of length 2. All compositions were formed with tasks at degradation level \textit{medium} (M).

\section{Comparison to StyleGAN-based methods}


\begin{figure*}[th!]
    \includegraphics[width=\linewidth]{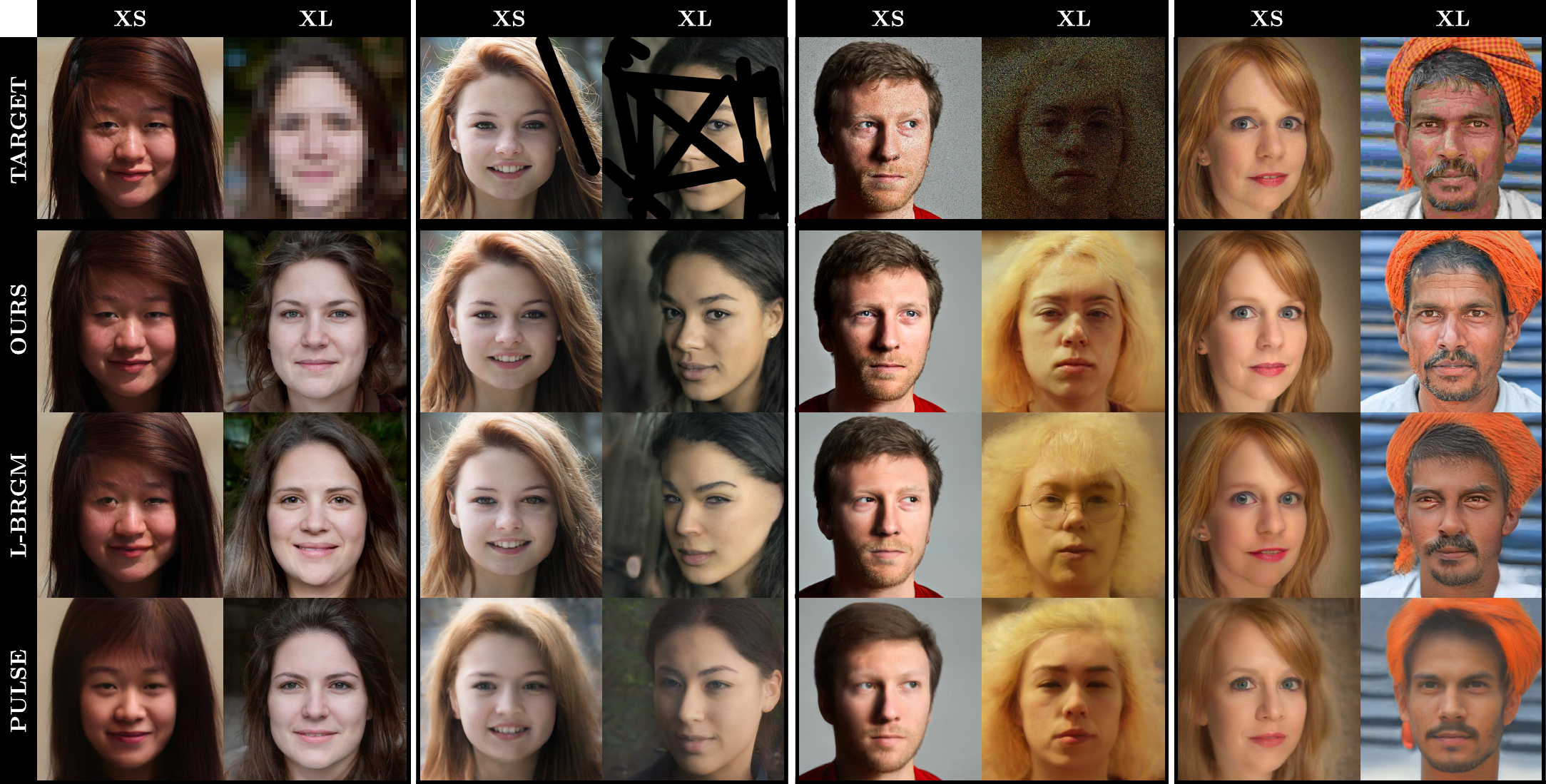}
    \caption{Randomly selected qualitative restoration results on each independent task, for both \textit{extra-small} and \textit{extra-large} levels of degradation. Top to bottom: (degraded) target, ours, L-BRGM~\cite{l-brgm}, and PULSE~\cite{pulse}. Left to right: upsampling, inpainting, denoising, and deartifacting. Our method shows better robustness to variations in levels of degradation. See supp. for more results.}
    \label{fig:single-degradations}
\end{figure*}

\begin{table}[t!]
\centering
\footnotesize
\begin{tabular}{@{}cccccccccccc@{}}
\toprule
& 
\multicolumn{3}{c}{Accur. \scriptsize(LPIPS) \small$\downarrow$} && 
\multicolumn{3}{c}{Fidelity  \scriptsize{(LPIPS)} \small$\downarrow$} && 
\multicolumn{3}{c}{Realism \scriptsize(pFID) \small$\downarrow$} \\
& 
\scriptsize{\textbf{PULS}} & 
\scriptsize{\textbf{L-BRG}} & 
\scriptsize{\textbf{OURS}} & \: & 
\scriptsize{\textbf{PULS}} & 
\scriptsize{\textbf{L-BRG}} & 
\scriptsize{\textbf{OURS}} & \: & 
\scriptsize{\textbf{PULS}} & 
\scriptsize{\textbf{L-BRG}} & 
\scriptsize{\textbf{OURS}}\\ 
\midrule
\multicolumn{12}{l}{\textbf{Upsampling} \scriptsize{(bilinear, bicubic or Lanczos)}}\\
\scriptsize\textit{XS} & .493 & \cellcolor{best}{.407} & \cellcolor{second}{.414} & & .432 & \cellcolor{best}{.295} & \cellcolor{second}{.313} & & 44.5 & \cellcolor{second}{23.6} & \cellcolor{best}{17.0}\\
\scriptsize\textit{S} & .492 & \cellcolor{best}{.412} & \cellcolor{second}{.449} & & .353 & \cellcolor{best}{.140} & \cellcolor{second}{.239} & & 34.3 & \cellcolor{second}{25.5} & \cellcolor{best}{22.0}\\
\scriptsize\textit{M} & .495 & \cellcolor{best}{.458} & \cellcolor{second}{.472} & & .261 & \cellcolor{best}{.124} & \cellcolor{second}{.172} & & \cellcolor{second}{29.3} & 35.4 & \cellcolor{best}{22.3}\\
\scriptsize\textit{L} & .501 & \cellcolor{best}{.487} & \cellcolor{second}{.490} & & .185 & \cellcolor{second}{.129} & \cellcolor{best}{.127} & & \cellcolor{second}{21.9} & 26.0 & \cellcolor{best}{20.9}\\
\scriptsize\textit{XL} & \cellcolor{second}{.512} & \cellcolor{best}{.506} & .514 & & \cellcolor{best}{.083} & .095 & \cellcolor{second}{.090} & & 24.9 & \cellcolor{second}{21.3} & \cellcolor{best}{21.3}\\
\midrule
\multicolumn{12}{l}{\textbf{Denoising} \scriptsize{(clamped Poisson and Bernoulli mixture)}}\\
\scriptsize\textit{XS} & .501 & \cellcolor{second}{.440} & \cellcolor{best}{.425} & & .275 & \cellcolor{best}{.152} & \cellcolor{second}{.156} & & 56.1 & \cellcolor{second}{27.2} & \cellcolor{best}{18.5}\\
\scriptsize\textit{S} & .499 & \cellcolor{second}{.450} & \cellcolor{best}{.434} & & .252 & \cellcolor{best}{.138} & \cellcolor{second}{.140} & & 53.7 & \cellcolor{second}{28.6} & \cellcolor{best}{19.1}\\
\scriptsize\textit{M} & .500 & \cellcolor{second}{.465} & \cellcolor{best}{.446} & & .224 & \cellcolor{second}{.155} & \cellcolor{best}{.130} & & 54.5 & \cellcolor{second}{22.1} & \cellcolor{best}{19.8}\\
\scriptsize\textit{L} & .501 & \cellcolor{second}{.481} & \cellcolor{best}{.457} & & .185 & \cellcolor{second}{.138} & \cellcolor{best}{.110} & & 56.4 & \cellcolor{second}{24.6} & \cellcolor{best}{19.2}\\
\scriptsize\textit{XL} & \cellcolor{second}{.504} & .511 & \cellcolor{best}{.474} & & .134 & \cellcolor{second}{.110} & \cellcolor{best}{.084} & & 49.4 & \cellcolor{second}{25.1} & \cellcolor{best}{17.9}\\
\midrule
\multicolumn{12}{l}{\textbf{Deartifacting} \scriptsize{(JPEG compression)}}\\
\scriptsize\textit{XS} & .498 & \cellcolor{second}{.442} & \cellcolor{best}{.432} & & .404 & \cellcolor{best}{.341} & \cellcolor{second}{.349} & & 52.3 & \cellcolor{second}{26.3} & \cellcolor{best}{14.8}\\
\scriptsize\textit{S} & .497 & \cellcolor{second}{.448} & \cellcolor{best}{.437} & & .398 & \cellcolor{second}{.352} & \cellcolor{best}{.350} & & 49.6 & \cellcolor{second}{22.4} & \cellcolor{best}{15.4}\\
\scriptsize\textit{M} & .498 & \cellcolor{second}{.461} & \cellcolor{best}{.445} & & .413 & \cellcolor{second}{.357} & \cellcolor{best}{.357} & & 33.2 & \cellcolor{second}{24.1} & \cellcolor{best}{15.4}\\
\scriptsize\textit{L} & .500 & \cellcolor{second}{.475} & \cellcolor{best}{.460} & & .395 & \cellcolor{best}{.367} & \cellcolor{second}{.374} & & 46.9 & \cellcolor{second}{25.2} & \cellcolor{best}{16.0}\\
\scriptsize\textit{XL} & .508 & \cellcolor{second}{.503} & \cellcolor{best}{.490} & & .427 & \cellcolor{second}{.418} & \cellcolor{best}{.412} & & 30.8 & \cellcolor{second}{22.1} & \cellcolor{best}{18.7}\\
\midrule
\multicolumn{12}{l}{\textbf{Inpainting} \scriptsize{(random strokes)}}\\
\scriptsize\textit{XS} & .498 & \cellcolor{second}{.409} & \cellcolor{best}{.378} & & .464 & \cellcolor{second}{.374} & \cellcolor{best}{.348} & & 46.9 & \cellcolor{second}{24.4} & \cellcolor{best}{12.9}\\
\scriptsize\textit{S} & .501 & \cellcolor{second}{.425} & \cellcolor{best}{.387} & & .356 & \cellcolor{second}{.287} & \cellcolor{best}{.264} & & 42.3 & \cellcolor{second}{27.2} & \cellcolor{best}{14.2}\\
\scriptsize\textit{M} & .509 & \cellcolor{second}{.438} & \cellcolor{best}{.396} & & .283 & \cellcolor{second}{.227} & \cellcolor{best}{.206} & & 38.5 & \cellcolor{second}{30.1} & \cellcolor{best}{14.5}\\
\scriptsize\textit{L} & .513 & \cellcolor{second}{.452} & \cellcolor{best}{.409} & & .231 & \cellcolor{second}{.184} & \cellcolor{best}{.163} & & \cellcolor{second}{32.6} & 33.1 & \cellcolor{best}{15.3}\\
\scriptsize\textit{XL} & .524 & \cellcolor{second}{.460} & \cellcolor{best}{.422} & & .187 & \cellcolor{second}{.157} & \cellcolor{best}{.132} & & 36.2 & \cellcolor{second}{25.2} & \cellcolor{best}{15.9}\\

\bottomrule
\end{tabular}
\caption{Quantitative comparison on individual tasks against baselines (``PULS''~\cite{pulse} and ``L-BRG''~\cite{l-brgm}). Baselines are optimized for accuracy on each row separately, while ours uses the same set of hyperparameters across all. Despite this clear handicap, our method is either on par or outperforms baselines. Results color-coded as \colorbox{best}{best} and \colorbox{second}{second best}.}
\label{tbl:degradation-levels}
\end{table}

\begin{table}[ht]
\centering
\footnotesize
\begin{tabular}{@{}rccccccccccc@{}}
\toprule
& 
\multicolumn{3}{c}{Accur. \scriptsize(LPIPS) \small$\downarrow$} && 
\multicolumn{3}{c}{Fidelity  \scriptsize{(LPIPS)} \small$\downarrow$} && 
\multicolumn{3}{c}{Realism \scriptsize(pFID) \small$\downarrow$} \\
& 
\scriptsize{\textbf{PULS}} & 
\scriptsize{\textbf{L-BRG}} & 
\scriptsize{\textbf{OURS}} & \: & 
\scriptsize{\textbf{PULS}} & 
\scriptsize{\textbf{L-BRG}} & 
\scriptsize{\textbf{OURS}} & \: & 
\scriptsize{\textbf{PULS}} & 
\scriptsize{\textbf{L-BRG}} & 
\scriptsize{\textbf{OURS}}\\ 
\midrule
\multicolumn{12}{l}{\textbf{2 degradations}} \\
\scriptsize\textit{NA} & .517 & \cellcolor{second}{.485} & \cellcolor{best}{.459} & & .328 & \cellcolor{second}{.301} & \cellcolor{best}{.290} & & 43.4 & \cellcolor{second}{24.2} & \cellcolor{best}{17.3}\\
\scriptsize\textit{AP} & .511 & \cellcolor{second}{.478} & \cellcolor{best}{.457} & & .270 & \cellcolor{second}{.231} & \cellcolor{best}{.204} & & 29.7 & \cellcolor{second}{17.6} & \cellcolor{best}{17.0}\\
\scriptsize\textit{UA} & \cellcolor{second}{.510} & .518 & \cellcolor{best}{.508} & & \cellcolor{second}{.307} & .348 & \cellcolor{best}{.287} & & 23.7 & \cellcolor{second}{20.5} & \cellcolor{best}{19.7}\\
\scriptsize\textit{NP} & .511 & \cellcolor{second}{.480} & \cellcolor{best}{.458} & & .125 & \cellcolor{second}{.079} & \cellcolor{best}{.062} & & 47.0 & \cellcolor{second}{20.9} & \cellcolor{best}{19.2}\\
\scriptsize\textit{UN} & \cellcolor{best}{.501} & .519 & \cellcolor{second}{.511} & & .178 & \cellcolor{best}{.149} & \cellcolor{second}{.153} & & 33.4 & \cellcolor{second}{26.2} & \cellcolor{best}{21.1}\\
\scriptsize\textit{UP} & .510 & \cellcolor{best}{.478} & \cellcolor{second}{.485} & & .140 & \cellcolor{best}{.061} & \cellcolor{second}{.089} & & \cellcolor{second}{23.9} & 35.3 & \cellcolor{best}{20.7}\\
\midrule
\multicolumn{12}{l}{\textbf{3 degradations}} \\
\scriptsize\textit{UNP} & \cellcolor{second}{.510} & .526 & \cellcolor{best}{.507} & & .086 & \cellcolor{second}{.062} & \cellcolor{best}{.051} & & 28.5 & \cellcolor{second}{22.1} & \cellcolor{best}{20.1}\\
\scriptsize\textit{UAP} & .525 & \cellcolor{second}{.523} & \cellcolor{best}{.513} & & .205 & \cellcolor{second}{.154} & \cellcolor{best}{.119} & & 23.0 & \cellcolor{best}{18.4} & \cellcolor{second}{20.5}\\
\scriptsize\textit{UNA} & \cellcolor{best}{.521} & .535 & \cellcolor{second}{.533} & & \cellcolor{best}{.265} & .310 & \cellcolor{second}{.290} & & 26.2 & \cellcolor{best}{20.7} & \cellcolor{second}{22.8}\\
\scriptsize\textit{NAP} & .526 & \cellcolor{second}{.502} & \cellcolor{best}{.470} & & .210 & \cellcolor{second}{.197} & \cellcolor{best}{.160} & & 38.7 & \cellcolor{best}{18.4} & \cellcolor{second}{18.5}\\
\midrule
\multicolumn{12}{l}{\textbf{4 degradations}} \\
\scriptsize\textit{UNAP} & \cellcolor{second}{.533} & .546 & \cellcolor{best}{.525} & & .192 & \cellcolor{second}{.177} & \cellcolor{best}{.131} & & 25.5 & \cellcolor{best}{21.1} & \cellcolor{second}{21.8}\\

\bottomrule
\end{tabular}
\caption{Quantitative comparison on composed tasks (at \textit{medium} levels) against baselines (``PULS''~\cite{pulse} and ``L-BRG''~\cite{l-brgm}). Baselines are optimized for accuracy on each row separately, while ours uses the same set of hyperparameters across all. Despite this clear handicap, our method is either on par or outperforms baselines. 
Acronyms indicate restorations: \textbf{U}psampling, de\textbf{N}oising, de\textbf{A}rtifacting, and/or in\textbf{P}ainting. Results color-coded as \colorbox{best}{best} and \colorbox{second}{second best}.}
\label{tbl:compositions}
\end{table}

\subsection{Models, datasets, and metrics}
\label{sec:models-datasets-metrics}

The PyTorch implementation~\cite{stylegan2-ada-pytorch} of StyleGAN2-ADA~\cite{stylegan2-ada} is used in all experiments. In particular, we use the model pretrained on the FFHQ dataset~\cite{stylegan} at $1024 \times 1024$ resolution. Since the model is trained on the entire FFHQ dataset, we gathered an additional 100 test images, dubbed ``FFHQ extras'' (FFHQ-X), using the same alignment script. Overlap with FFHQ was avoided by filtering by date (see supp.). 

Performance was measured via accuracy, realism, and fidelity metrics. 
Accuracy was measured with the LPIPS~\cite{lpips} between the predictions and the ground truth images. 
Realism was measured using the PatchFID~\cite{anyres} (pFID): FID~\cite{fid} measured on image patches. More specifically, we extracted 1000 random crops of size $128 \times 128$ per image, computing the FID of all crops from predictions to all crops from our FFHQ-X dataset. 
Finally, fidelity between degraded predictions and targets was measured using the LPIPS; this was done by degrading predictions with the \emph{true} degradation to ignore the effect of differentiable approximations.

\subsection{Baselines}
\label{sec:baselines}

Our method is compared on FFHQ-X against two unsupervised StyleGAN image restoration techniques, PULSE~\cite{pulse} and the current state of the art L-BRGM~\cite{l-brgm}. The same pretrained network is used for all approaches. 

\paragraph{PULSE~\cite{pulse}} We adapted PULSE to use StyleGAN2-ADA and to tasks other than upsampling by increasing the number of steps from $100$ to $250$ and adjusting the learning rate to $0.8$ (manual inspection on validation data) to increase its robustness to novel tasks. 

\paragraph{L-BRGM~\cite{l-brgm}} We adjust its learning rate $\lambda_\mathrm{lr} = 0.05$ by manual inspection. L-BRGM uses best-of-$k$ random initialization; because initialization is difficult under high degradations, we doubled the number of samples from $100$ to $200$ and added a 0.8 truncation~\cite{stylegan} to avoid outliers. L-BRGM also performs early stopping by inspecting the ground truth; for a fair comparison, we instead fixed the number of steps to $500$ as in BRGM~\cite{brgm}. 

\paragraph{Hyperparameter optimization for baselines} We optimize the accuracy metric for the baselines hyperparameters for each (task, level) pair individually \emph{directly on the FFHQ-X test set} (see supp. for more details). In addition, baselines employ the ground truth Bernouilli mask (see \cref{sec:individual-degradations}) in their differentiable approximation of \cref{eq:noise-model}. Doing so was necessary for good performance. 




\paragraph{Our method} We adjust hyperameters (learning rate and number of iterations, see \cref{pseudocode}) once on validation data, then run our benchmark on the test set with \emph{no per-task or per-level hyperparameter adjustment}: our method uses the same hyperparameters across all tasks and levels. Our method is not provided the Bernouilli mask. The total number of steps (450) is less than the 500 used for L-BRGM.

\subsection{Experimental results}
\label{sec:experiments}

\paragraph{Robustness to degradation levels} 
\label{expr:robustness}
Our benchmark is run on our FFQH-X test set (\cref{sec:models-datasets-metrics}) on all five levels of degradation for each individual task, and quantitative results are presented in \cref{tbl:degradation-levels}. Recall from \cref{sec:baselines} that our method, which employs a single set of hyperparameters across all experiments, is pitted against baselines with hyperparameters optimized for each task and level individually. Despite this obvious impediment, our proposed method outperforms baselines in the vast majority of scenarios. In particular, it reaches better accuracy in the majority of tasks against baselines \emph{optimized for accuracy}, and realism is strictly better. \Cref{fig:single-degradations} illustrates the improved visual quality obtained with our method. 

\begin{figure*}[th]
    \includegraphics[width=\linewidth]{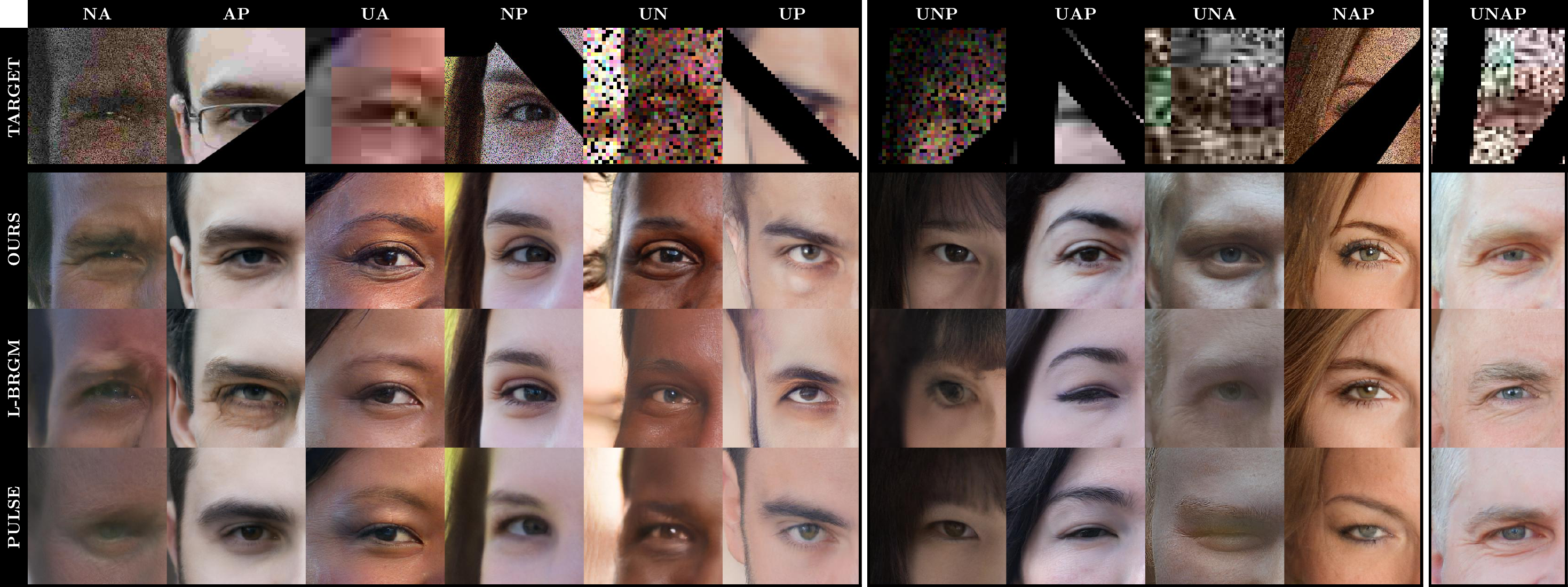}
    \caption{Qualitative comparisons on all possible compositions of the four tasks given \textit{medium} levels of degradation. Samples are chosen randomly, and acronyms indicate the degradations: \textbf{U}psampling, in\textbf{P}ainting, de\textbf{N}oising and/or de\textbf{A}rtifacting (\cref{expr:compositions}).}
    \label{fig:composed-degradations}
\end{figure*}

\paragraph{Robustness to compositions} 
\label{expr:compositions}
Robustness to compositions of degradations was evaluated by forming all possible subsequences of degradations (order in \cref{eq:composition-order}) at \textit{medium} level. Quantitative results are aggregated in \cref{tbl:compositions}, while \cref{fig:composed-degradations} presents qualitative examples. Our results show that both high fidelity and high realism are maintained with our method, while the baselines struggle under varied degradations.

\begin{table}[t!]
\centering
\footnotesize
\begin{tabular}{
m{1.4cm}<{\centering}
m{1.4cm}<{\centering}
m{0.1cm}<{\centering}
m{1.4cm}<{\centering}
m{1.4cm}<{\centering}
}
\toprule
\multicolumn{2}{c}{Accur. \scriptsize(LPIPS) \small$\downarrow$} & &
\multicolumn{2}{c}{Realism \scriptsize(pFID) \small$\downarrow$} \\
\scriptsize{\textbf{DDRM}} & 
\scriptsize{\textbf{OURS}} & &
\scriptsize{\textbf{DDRM}} & 
\scriptsize{\textbf{OURS}} \\ 
\midrule
\multicolumn{5}{l}{\textbf{Exact Upsampling} \scriptsize(known average pooling)} \\
\cellcolor{best}{0.412} & 0.443 & & 44.75 & \cellcolor{best}{18.92} \\
\multicolumn{5}{l}{\textbf{Inexact Upsampling} \scriptsize{(unknown nearest-neighbor)}}\\
0.559 & \cellcolor{best}{0.539} & & 67.06 & \cellcolor{best}{32.81} \\
%
\midrule
\multicolumn{5}{l}{\textbf{Linear Denoising} \scriptsize{(additive Gaussian noise)}}\\
\cellcolor{best}{0.336} & 0.452 & & 43.28 & \cellcolor{best}{19.25} \\ 
\multicolumn{5}{l}{\textbf{Nonlinear Denoising} \scriptsize{(clamped additive Gaussian noise)}}\\
0.530 & \cellcolor{best}{0.450} & & 156.1 & \cellcolor{best}{19.06} \\
\midrule
\multicolumn{5}{l}{\textbf{Inpainting} \scriptsize{(masking random strokes)}}\\
\cellcolor{best}{0.289} & 0.301 & & 26.12 & \cellcolor{best}{25.71} \\
\multicolumn{5}{l}{\textbf{Uncropping} \scriptsize{(top left corner unmasked)}}\\
0.468 & \cellcolor{best}{0.418} & & 34.33 & \cellcolor{best}{17.645} \\
\bottomrule
\end{tabular}
\caption{Quantitative comparison to DDRM~\cite{ddrm} on accuracy and realism (see \cref{sec:comparison-diffusion}). Results are color-coded as \colorbox{best}{best}.}
\label{tbl:diffusion-results}
\end{table}


\begin{figure}[t]
    \centering
    \includegraphics[width=\linewidth]{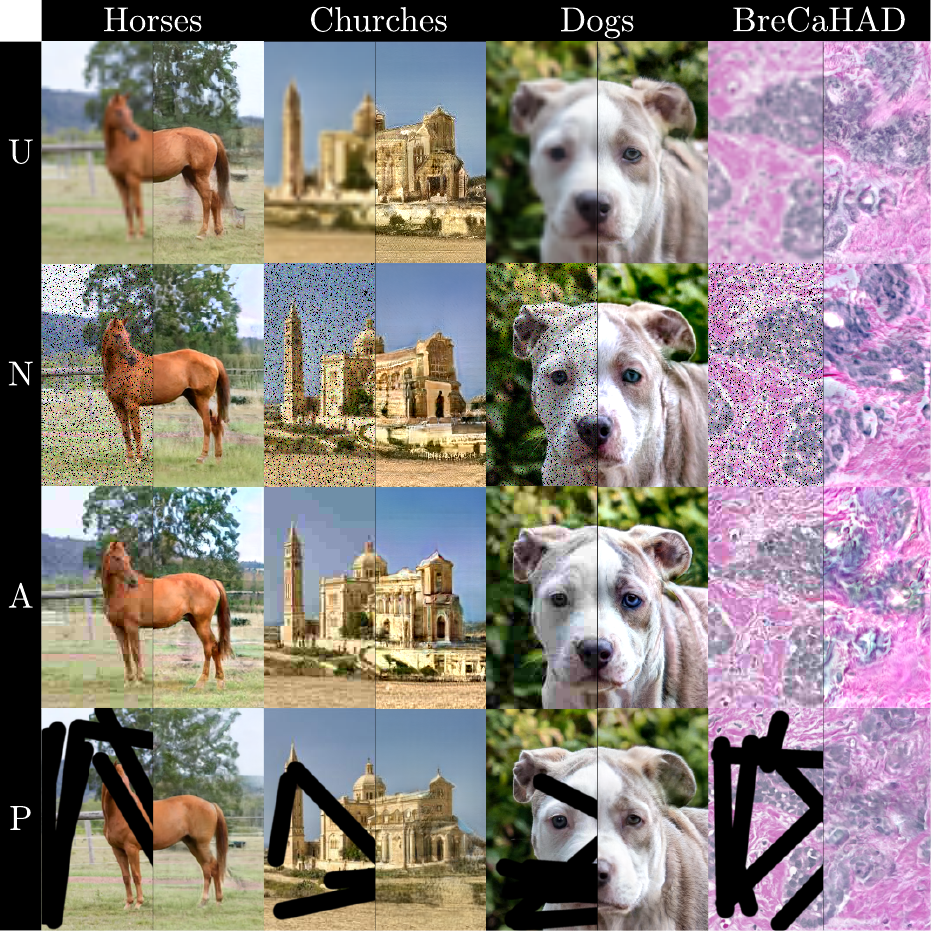}
    \caption{Results on LSUN horses and churches~\cite{yu15lsun}, AFHQ-v2 Dogs~\cite{choi2020starganv2} and BreCaHAD~\cite{aksac2019brecahad} at resolution $256 \times 256$ for small upsampling (U), small denoising (N), medium deartifacting (A), and medium inpainting (P); the left/right halves are the target/restored images. See supp. for details and more results.}
    \label{fig:other-datasets}
\end{figure}

\section{Comparison to diffusion models}
\label{sec:comparison-diffusion}

We compare our approach to the recent ``Denoising diffusion restoration model'' (DDRM)~\cite{ddrm}, which also does not require problem-specific supervised training and leverages a pre-trained (DDPM~\cite{ddpm}) model. Importantly, DDRM is designed only for \emph{linear} inverse problems with additive Gaussian noise, where the degradation model is \emph{perfectly known}. Therefore, for each of the upsampling, denoising and inpainting tasks, we experiment with both: 1) a linear inverse problem with known parameters; and 2) an adjusted version which is either non-linear, where the extract degradation model is unknown, or a more extreme version of the problem. Specifically, we test: average pooling (known, linear) vs nearest-neighbor (unknown) $32\times$ (XL) upsampling; additive Gaussian noise (linear) vs clamped (to $[0, 255]$) additive Gaussian noise (non-linear) with $\sigma^2=0.8$ (XL) for both; and masking strokes (at XL level) vs uncropping the top-left corner (XXL). DDRM is available on CelebA (not FFHQ) at $256 \times 256$, so we use the corresponding StyleGAN2-ADA pretrained model. Due to the reduced resolution, pFID is measured on 250 patches of size $64 \times 64$. 

Interestingly, we find that DDRM reproduces the targets almost perfectly (fidelity is near zero, see supp. for details and qualitative results), to the point where it overfits to any deviation between its degradation model and the true degradation. \Cref{tbl:diffusion-results} presents the accuracy and realism metrics for all the aforementioned tasks.  While the accuracy favors DDRM for the tasks which perfectly match its assumptions, our approach outperforms it when this is not the case. In addition, our method provides significantly better realism across \emph{all} scenarios. Overall, our approach yields high resolution results with more detail, can adapt to non-linear tasks, and it can also cope with larger degradations. 
\section{Discussion}
\label{sec:discussion}

\paragraph{Limitations}
Our method is not restricted to faces and works with any pretrained StyleGAN, as demonstrated in \cref{fig:other-datasets}. However, it is inherently limited to the domain learned by the GAN. Training StyleGAN on a large unstructured dataset like ImageNet is difficult but recent papers~\cite{stylegan2-distillation,stylegan-xl,gigagan} have attained some success---applying our method on some of these large-scale models is a promising research avenue. Our method also faces the same ethical considerations as the GAN it relies on. Finally, as with other methods~\cite{pulse,l-brgm,ddrm}, we also require knowledge of an (approximate) degradation function.  

\paragraph{Conclusion}
This paper presents a method which makes StyleGAN-based image restoration robust to both variability in degradation levels and to compositions of different degradations. Our proposed method relies on a conservative optimization procedure over a progressively richer latent space and avoids regularization terms altogether. Using a single set of hyperparameters, we obtain competitive and even state-of-the-art results on several challenging scenarios when compared to baselines that are optimized for each task/level individually. 

\paragraph{Acknowledgements} {\small This research was supported by NSERC grant RGPIN-2020-04799, a MITACS Globalink internship to Y. Poirier-Ginter, and by the Digital Research Alliance Canada.}

{\small
\bibliographystyle{ieee_fullname}
\bibliography{egbib}
}

\end{document}